\documentclass[letterpaper, 10 pt, conference]{ieeeconf}  % Comment this line out if you need a4paper

\IEEEoverridecommandlockouts                              % This command is only needed if 
\usepackage{epsfig} % for postscript graphics files
\usepackage{mathptmx} % assumes new font selection scheme installed
\usepackage{times} % assumes new font selection scheme installed
\usepackage{amssymb}  % assumes amsmath package installed
\usepackage{graphicx}
\usepackage{algorithm}

\usepackage{xcolor}
\usepackage{subcaption}

\usepackage{amsmath}
\usepackage{booktabs}
\usepackage{multirow}
\usepackage{hyperref}
\newcommand{\tpvqa}{TPVQA}

\title{\LARGE \bf
Grounding Classical Task Planners via Vision-Language Models
}

\author{Xiaohan Zhang$^1$, Yan Ding$^1$, Saeid Amiri$^1$, Hao Yang$^2$, Andy Kaminski$^2$, Chad Esselink$^2$, Shiqi Zhang$^1$% <-this % stops a space
\thanks{$^1$~Department of Computer Science, SUNY Binghamton}
\thanks{$^2$~Ford Motor Company}
}

\begin{document}

\maketitle
\thispagestyle{empty}
\pagestyle{empty}

%%%%%%%%%%%%%%%%%%%%%%%%%%%%%%%%%%%%%%%%%%%%%%%%%%%%%%%%%%%%%%%%%%%%%%%%%%%%%%%%
\begin{abstract}
% Classical planning systems are good at leveraging rule-based human knowledge to compute correct plans but suffer from the strong assumptions of perfect perception and action executions.
% One approach to address these issues is to link the symbolic states and actions from classical planners to the robot's sensory observations, thereby closing the perception-action loop.
% In this paper, we propose a vision-based symbolic planning framework, called \tpvqa, that leverages Vision-Language Models~(VLMs) to monitor action failures and verify action affordances for reliable and successful plan execution.
% Experimental results demonstrate that \tpvqa ~outperforms competitive baselines from the literature, achieving the highest task completion rate.

Classical planning systems have shown great advances in utilizing rule-based human knowledge to compute accurate plans for service robots, but they face challenges due to the strong assumptions of perfect perception and action executions. 
To tackle these challenges, one solution is to connect the symbolic states and actions generated by classical planners to the robot's sensory observations, thus closing the perception-action loop. 
This research proposes a visually-grounded planning framework, named \tpvqa, which leverages Vision-Language Models~(VLMs) to detect action failures and verify action affordances towards enabling successful plan execution. 
Results from quantitative experiments show that \tpvqa ~surpasses competitive baselines from previous studies in task completion rate.

\end{abstract}

\section{Introduction}
Classical planning frameworks such as those defined by Planning Domain Definition Language~(PDDL) and Answer Set Programming~(ASP) have been extensively utilized in planning and reasoning robot actions for long-horizon tasks~\cite{ghallab2016automated}.
% However, these frameworks often make certain assumptions that may not hold true in real-world scenarios.
% Specifically, they assume that the robot possesses complete knowledge of the world model by being able to perfectly perceive world states in advance. 
% Additionally, they assume that actions cause deterministic state transitions. 
Those classical planning systems are good at leveraging rule-based human knowledge to compute correct plans but suffer from the strong assumptions of perfect perception and action executions.
For example, if the world model includes an apple on a table, classical planners assume that the robot will always locate the apple after reaching the table's location, and picking up the apple will deterministically result in it being in the robot's hand. 
These assumptions fail to consider dynamically changing environments and uncertain action outcomes, rendering it impractical for the robot to complete tasks by simply following computed plans in the real world.

\begin{figure}
\begin{center}
    \includegraphics[width=0.45\textwidth]{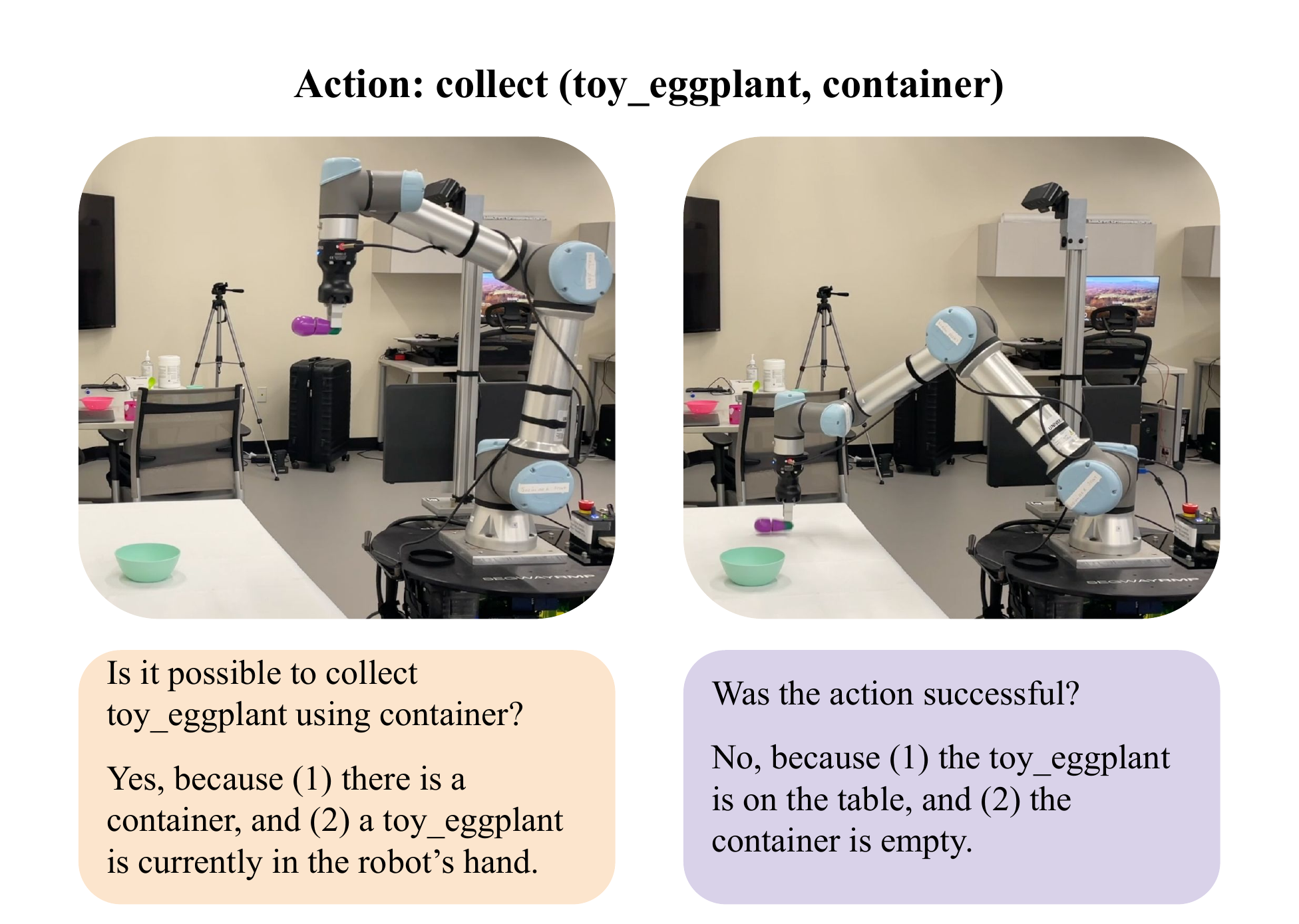}
    \caption{
    % \shiqi{The image backgrounds are a bit cluttered. Since it's a workshop submission, maybe we can go with what we have now.}
    To monitor plan executions on robots, it is required to answer the following questions: 1) \textit{``Is it feasible to perform a particular action in the current state of the world?''} and 2) \textit{``Was the action successfully executed, resulting in world transitions to the desired state?''}
    }
    \vspace{-2em}
    \label{fig:intro}
\end{center}
\end{figure}

\begin{figure*}
\begin{center}
    \includegraphics[width=0.95\textwidth]{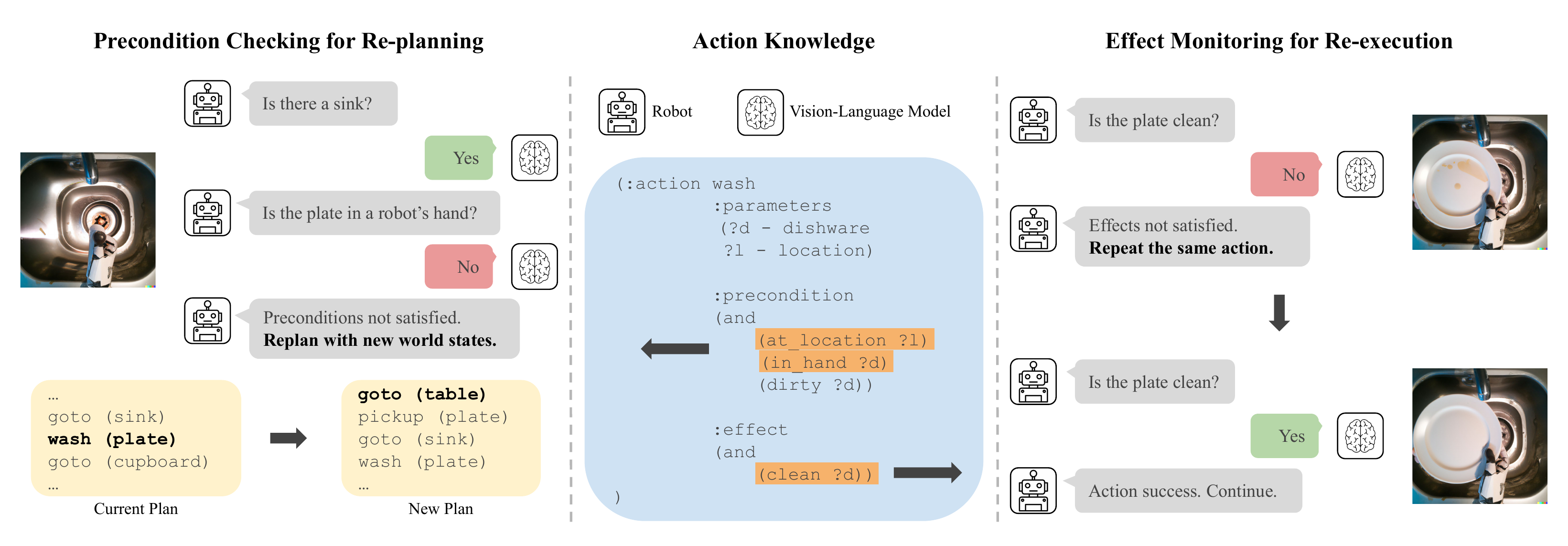}
    \caption{Overview of \tpvqa. ~By simply querying the robot's current observation against the action knowledge~(i.e., effects and preconditions) as Visual Question Answering~(VQA) tasks, \tpvqa ~can trigger the robot to repeat an unsuccessful action, or call the classical planner to generate a new valid plan using an updated world state.
    }
    \vspace{-2em}
    \label{fig:overview}
\end{center}
\end{figure*}

% To enable a robot to successfully complete tasks, one approach is to link the symbolic states and actions from classical planners to the robot's sensory observations, thereby closing the perception-action loop~(Fig.~\ref{fig:intro}).
To enable successful plan executions, classical planning systems are frequently accompanied by a plan monitoring system for linking the symbolic states and actions to robot sensory observations, where significant engineering efforts are needed. 
Fig.~\ref{fig:intro} illustrates the role of a plan monitoring system.
Given the natural connection between planning symbols and human language, this paper investigates how pre-trained Vision-Language Models~(VLMs) can assist the robot in realizing symbolic plans generated by classical planners, while avoiding the engineering efforts of checking the outcomes of each action. 
Specifically, we propose a vision-based symbolic planning framework, called \tpvqa, that leverages VLMs to detect action failures and verify action affordances towards successful plan execution~(Fig.~\ref{fig:overview}). 
We take the advantage of the domain knowledge encoded in classical planners, including the actions defined by their effects and preconditions.
By simply querying current observations against the action knowledge, similar to applying VLMs to Visual Question Answering~(VQA) tasks, \tpvqa ~can trigger the robot to repeat an unsuccessful action or call the symbolic planner to generate a new valid plan.
% \shiqi{I am not sure how to interpret ``or even...'' here. If there's a switch, let's be clear about the condition.}

We conducted quantitative evaluations of \tpvqa~on an image dataset that consists of 20\% realistic photos taken from home environments.
The remaining images are augmented using diffusion models~\cite{ramesh2021zero}. 
Experimental results demonstrate that \tpvqa~outperforms competitive baselines from the literature, achieving the highest task completion rate. 
Furthermore, we present an illustrative trial of deploying \tpvqa~on real robot hardware to perform object rearrangement tasks.

\section{Related Work}
\subsection{Robot Planning with Classical Planners}

Classical planning algorithms have found widespread application in robot systems.
Recent classical planning systems designed for robotics commonly employ Planning Domain Description Language (PDDL) or Answer Set Programming (ASP) as the underlying action language for planners~\cite{jiang2019task, brewka2011answer, lifschitz2002answer, fox2003pddl2}.
Researchers have utilized classical planning algorithms for various robotic applications, including sequencing actions for a mobile robot on delivery tasks~\cite{zhang2015mobile}, reasoning about safe and efficient urban driving behaviors for autonomous vehicles~\cite{ding2020task, ding2022glad}, planning actions for a team of mobile robots~\cite{jiang2019multi}, learning to ground properties for unknown objects~\cite{ding2022learning}, and completing service tasks in open-world scenarios~\cite{ding2022robot}.
Task and Motion Planning (TAMP), a hierarchical planning framework that combines classical planning in discrete spaces and robot motion planning in continuous space, has also shown great advances in robot long-horizon planning~\cite{lagriffoul2018platform, kaelbling2013integrated}.
Most classical planning algorithms that are designed for robot planning do not consider perception.
Though some recent works have already shown that training vision-based models from robot sensory data can be effective in plan feasibility evaluation~\cite{zhu2021hierarchical, zhang2022visually, driess2020deep, driess2020deeph, wells2019learning, ding2023task}, their methods did not tightly bond with language symbols which are the state representations for classical planning systems. 
We, on the other hand, propose \tpvqa ~that uses VLMs to connect language~(in classical planner) to robot perception.
% \shiqi{Conclude this para with their limitations. Will come back to it after you complete this sec. }

\subsection{Pre-trained Vision-Language Models in Robotics}
\label{sec:related_vlm}
% Large Vision-Language Models have demonstrated impressive performance in various robot applications.
Existing research has shown that large Vision-Language Models like CLIP~\cite{radford2021learning}  can be used in the robotics domain such as semantic scene understanding~\cite{ha2022semantic}, effective open-ended agent learning~\cite{fan2022minedojo}, guiding robot navigation~\cite{shafiullah2022clipfields} and manipulation behaviors~\cite{shridhar2021cliport, moo2023arxiv}.
Two recent works that are the most related to us are SuccessVQA~\cite{du2023vision} and PaLM-E~\cite{driess2023palm}.
SuccessVQA has investigated how VLMs enable robots to detect action outcomes and model action rewards.
They also treat success/failure detections as VQA tasks, but did not consider affordances before action execution. 
PaLM-E is a large embodied VLM that is trained to predict robot action sequences as well as solve other downstream vision-language tasks. 
PaLM-E has demonstrated its effectiveness in both failure detection and affordance prediction.
Different from their work, \tpvqa ~uses classical planners for generating symbolic plans instead of solely relying on pre-trained models.
That is because classical planning techniques are designed to ensure the generated plans are sound and complete.
In addition, all the works that are listed in this section require additional training or fine-tuning VLMs in specific domains, but we study if and to what extent existing VLMs can help robot planning.

\section{Background}
In this section, we briefly summarize basic concepts in classical planning and Vision-Language Models, which serve as the two main building blocks of this research.
\subsection{The Classical Planning Problem}
Formally, the input of a planning problem $P$ is defined by a tuple $\langle S, s^{init}, S^{G}, A, f\rangle$. 
$S$ is a finite and discrete set of states used to describe the world's state (i.e., state space).
We assume a factored state space such that each state $s \in S$ is defined by the values of a fixed set of variables.
$s^{init} \in S$ is an initial world state.
$S^{G} \subset S$ is a set of goal states. $S^{G}$ are usually specified as a list of \emph{goal conditions}, all of which must hold in a goal state.
$A$ is a set of symbolic actions. Actions are defined by their preconditions and effects.
$f$ is the underlying state transition function. 
% $f$ takes the current state and an action as input and outputs the corresponding next state. 
State transitions are usually deterministic in classical planning problems but are not in real-world scenarios.
% \begin{itemize}
%     \item $S$ is a finite and discrete set of states used to describe the world's state (i.e., state space).
%     We assume a factored state space such that each state $s \in S$ is defined by the values of a fixed set of variables.
%     \item $s^{init} \in S$ is an initial world state.
%     \item $S^{G} \subset S$ is a set of goal states. $S^{G}$ are usually specified as a list of \emph{goal conditions}, all of which must hold in a goal state.
%     \item $A$ is a set of symbolic actions. Actions are defined by their preconditions and effects.
%     \item $f$ is the underlying state transition function. $f$ takes the current state and an action as input and outputs the corresponding next state. 
%     State transitions are usually deterministic in classical planning problems but are not in real-world scenarios.
% \end{itemize}
A solution to a classical planning problem $P$ is a symbolic plan $\pi$ in the form of $\langle a_1, a_2, \dots, a_N \rangle$, such that the preconditions of $a_1$ hold in $s^{init}$, the preconditions of $a_2$ hold in the state that results from applying $a_1$, and so on, with the goal conditions all holding in the state that results after applying $a_N$.  

\subsection{Vision-Language Models for VQA Tasks}
A general definition for VLMs is that they are models that combine both vision and language modalities.
Most VLMs require encoders for both vision and language so as to train joint feature embeddings.
One typical training strategy is by using Contrastive Learning~\cite{chen2020simple}.
Pre-trained VLMs have shown impressive capabilities in downstream tasks such as image captioning~\cite{vinyals2015show}, open-vocabulary object detection~\cite{minderer2022simple}, and visual question answering~\cite{antol2015vqa}.
In this work, we relate robot actions with VQA queries for grounding long-horizon planning. 
We use the ViLBERT model~\cite{lu2019vilbert} pre-trained on the VQA v2.0 dataset~\cite{goyal2017making} which is publicly available in the AllenNLP platform~\cite{gardner2018allennlp}.

\section{Method}
\label{sec:method}
This section presents our main contribution, \tpvqa, that leverages VLMs to detect action failures and verify action affordances for enabling successful plan executions.

\subsection{Precondition Checking for Re-planning}
Before every action execution, \tpvqa ~extracts knowledge about action preconditions from the planner's domain description.
For instance, action \texttt{place\_on($A$, $B$)} has preconditions of \texttt{in\_hand($A$)} and \texttt{near($B$)}, meaning that to place an object $A$ on top of object $B$, the robot should first grasp $A$ in hand and be located near object $B$.
Then, we simply convert each action precondition into a natural language query by using some manually defined templates, such as \textit{``Is object $A$ in a robot's hand?''} and \textit{``Is there an object $B$ in the image?''}
% The conversions are implemented by using some manually defined templates.\xiaohan{Adding details about templates here. Maybe an example.}
Paring each natural language query with the current observation from the robot's first-person view, we call the VLM to get answers indicating if the precondition is satisfied.

According to the results from the VLM, \tpvqa ~will update the current state information in the classical planning system.
Fig.~\ref{fig:overview} shows an example where the robot wants to \texttt{wash(plate)} but fails to detect ``plate in a robot's hand'' given the current image.
Because the classical planner always assumes perfect action executions, it will incorrectly believe all previous actions are successful and the current world state includes \texttt{in\_hand(plate)}.
As a result, \tpvqa ~will update the current state by removing \texttt{in\_hand(plate)}.
We provide the updated world state to the planner as the ``new'' initial state to re-generate a plan. 
In the above example, instead of \texttt{wash(plate)}, the robot will now take the action of \texttt{goto(table)} as it believes there is a plate on the table~(according to the domain knowledge provided in the planning problem description). 

\subsection{Effect Monitoring for Re-execution}
After every action execution, \tpvqa ~extracts knowledge about action effects from the planner's domain description. 
Similar to how \tpvqa ~asks about preconditions, it queries action effects by using the VLM. 
If the effects are not satisfied, the robot will repeat the same action until it gets positive feedback from the VLM so as to continue the next action.
Note that before re-trying each action, the robot will also need to check preconditions, because action failures frequently cause some preconditions to break in the real world. 
For example, failing to place an apple on the table might result in the apple falling to the ground, instead of still being in the robot's hand. 
\newline
\vspace{-1em}

\noindent\textbf{Remarks: }
A single action is usually defined by multiple preconditions and effects.
VLMs, especially for those that are not trained using domain-specific data, frequently produce inaccurate answers that cause disagreements among the given preconditions~(or effects).
For instance, the VLM might answer ``Yes'' to both \texttt{on(apple, table)} and \texttt{in\_hand(apple)} after the robot picks up an apple from the table.
In this paper, we query the VLM about all the listed effects~(preconditions), and determine to re-execute~(re-plan) if the majority of them are not satisfied.

\section{Experiments}
We conduct extensive experiments to evaluate the performance of \tpvqa ~comparing with baselines from the literature.
Our hypothesis is that \tpvqa ~produces the highest task completion rate because of its effectiveness in plan monitoring and online re-planning using perception.
In the experiment, we consider three everyday tasks that are ``clean\_dishes'', ``serve\_breakfast'', and ``eat\_apple''.
Task descriptions are constructed using PDDL and symbolic plans are generated using the \textsc{fast-downward} planner\footnote{See \url{https://www.fast-downward.org/} for the details on the \textsc{fast-downward} software. We use the implementation from \url{https://github.com/aibasel/downward}.}, as shown in TABLE~\ref{tab:tasks}.

\begin{table}[h!]
    \centering
    \scriptsize
    \begin{tabular}{l l l}
    \toprule 
    clean\_dishes             & serve\_breakfast        & eat\_apple\\
    \midrule
    Step 1: Find plate             &  Step 1: Find bread       &  Step 1: Find fridge\\
    Step 2: Pick up plate             &  Step 2: Pick up bread       &  Step 2: Open fridge\\
    Step 3: Find sink             &  Step 3: Find plate       &  Step 3: Find apple\\
    Step 4: Wash plate             &  Step 4: Place bread on plate      &  Step 4: Pick up apple\\
             &  Step 5: Find TV       &  Step 5: Find knife\\
             &  Step 6: Turn on TV      &  Step 6: Pick up knife\\
             &        &  Step 6: Cut into half apple\\    
    \bottomrule
    \end{tabular}
\caption{Symbolic plans computed for the tasks.}
\vspace{-2em}
\label{tab:tasks}
\end{table}

\subsection{Simulator with Diffusion Models}
To quantitatively evaluate the performance of \tpvqa ~in dealing with imperfect perception and uncertain action outcomes, we build a simulator using web-scale diffusion models.
\begin{figure}
\begin{center}
    \includegraphics[width=0.45\textwidth]{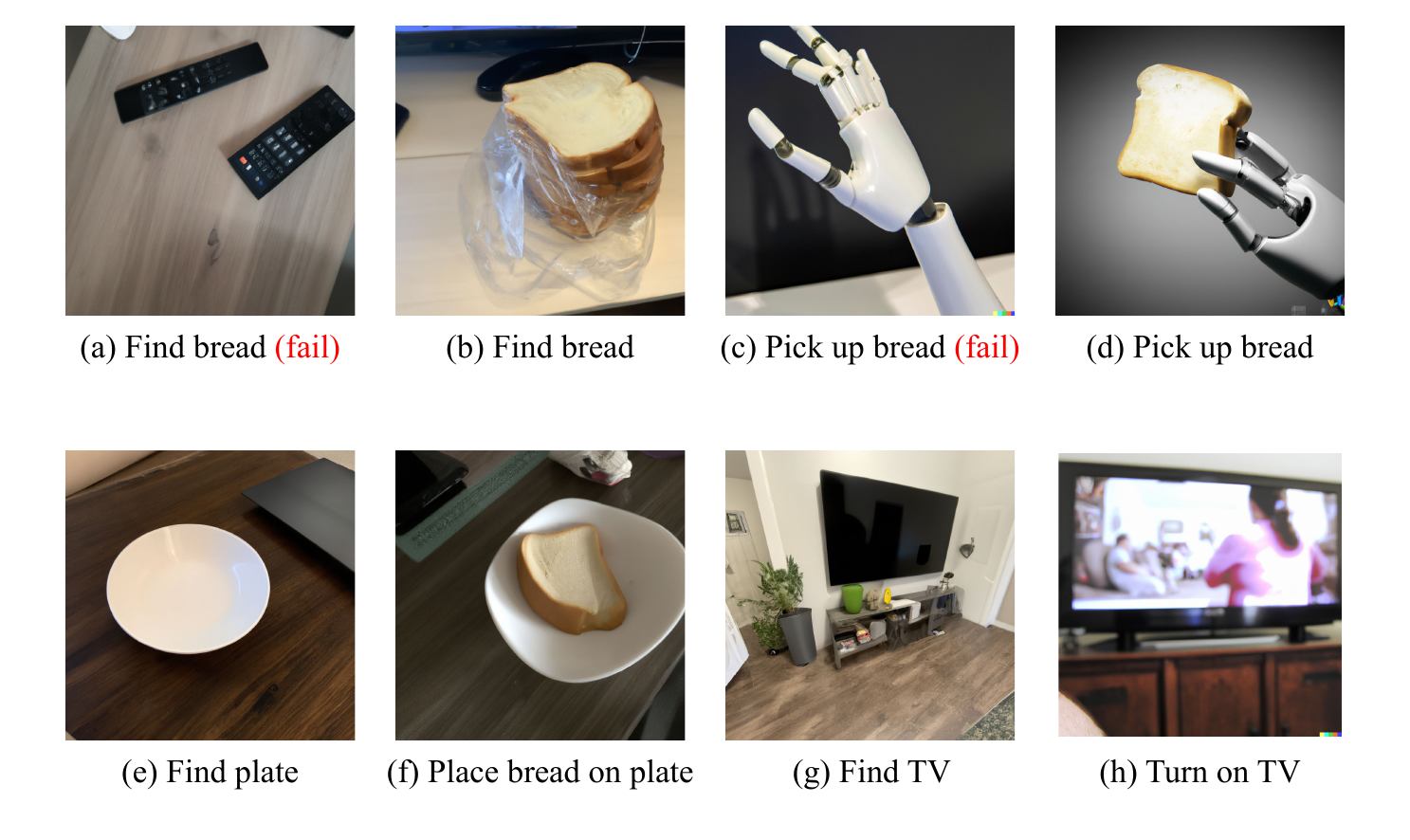}
    \caption{An example trial for ``serve\_breakfast'' task sampled from the simulator.
    % \shiqi{Somehow index the subfigures to be clear about the ordering.}
    }
    \vspace{-3em}
    \label{fig:dalle}
\end{center}
\end{figure}
We first took images from real environments and use the image variation API provided by DALL-E~\cite{ramesh2021zero} to augment the original dataset.
For a small portion of the actions for which real photos are difficult to get, such as a robot washing a plate, we manually design prompts as inputs to DALL-E. 
Each action is paired with 10 successful observations and 10 failed ones.
Overall, our image dataset consists of 20\% real photos, 60\% images from real photo variations, and 20\% images directly generated from text prompts.
Fig.~\ref{fig:dalle} shows example images from our dataset.

At each time step, an observation for the current action is sampled from the dataset.
We assume that there is a probability of 25\% that an action may fail which will result in a failed action observation. 
We also assume there is another 25\% chance that a failed action may cause changes to previous states.
For instance, when the robot fails on the action \texttt{cutintohalf(apple)}, there is a chance that the apple~(or the knife) is not in the robot's hand anymore.
To model this uncertainty, we re-sample one of the previous observations to let the robot estimate the current world state.
The robot needs to successfully execute all the actions so as to complete the task, leading the system to the desired goal state.
\subsection{Baselines}
\tpvqa ~is compared with the following four baselines:
\begin{itemize}
    \item EffectVQA: An ablative version of ours where VLMs are only used for action effect monitoring.
    \item TP: A task planning baseline without perception.
    \item PaLMEVQA: PaLM-E~\cite{driess2023palm} is robust to most of the vision-language downstream tasks.
    Among those, we are more interested in affordance prediction and failure detection.
    To this end, PaLMEVQA is a baseline that is designed with prompts provided in the original PaLM-E paper, which are \textit{``Is it possible to $<$action$>$ here?''} and \textit{``Was $<$action$>$ successful?''}
    \item SuccessVQA~\cite{du2023vision}: We use the same query provided in their paper, which is \textit{``Did the robot successfully $<$action$>$?''} SuccessVQA does not consider affordance.
\end{itemize}
Note that neither PaLM-E nor Flamingo~\cite{alayrac2022flamingo}~(as used in the original SuccessVQA paper) is open-sourced, so we use the same VLM as ours~\cite{lu2019vilbert} for implementing their corresponding baselines.
As discussed in Section~\ref{sec:related_vlm}, both PaLM-E and SuccessVQA are trained on robotics data, but all evaluations in this paper do not involve any altering for the VLM itself.
\subsection{Results}
\begin{figure}
\begin{center}
    \includegraphics[width=0.45\textwidth]{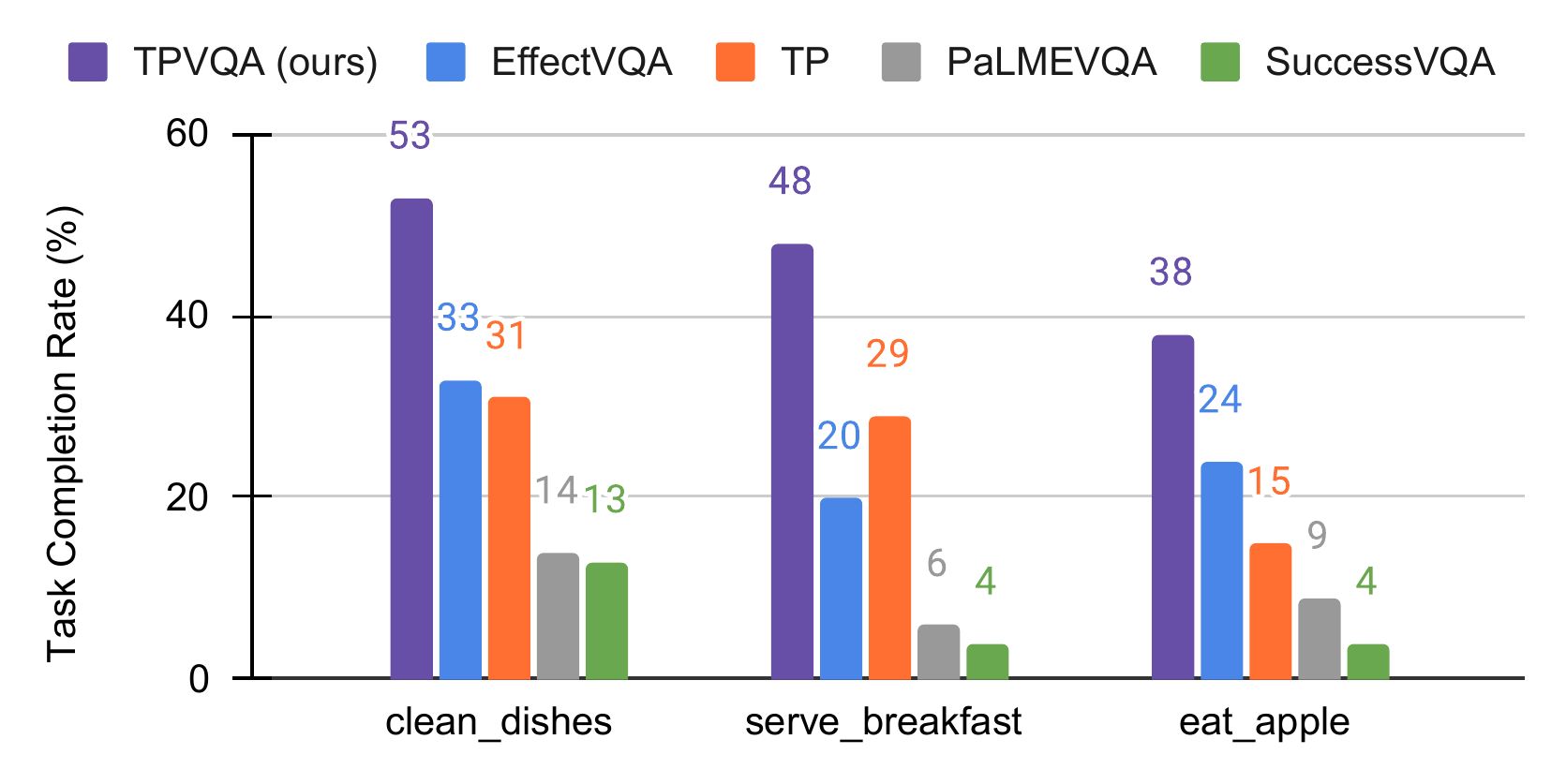}
    \caption{Task completion rates of \tpvqa ~and four baselines evaluated over three tasks.
    }
    \label{fig:main_result}
    \vspace{-2.5em}
\end{center}
\end{figure}

Fig.~\ref{fig:main_result} presents the main experimental results.
We observe that \tpvqa ~consistently outperforms baselines in task completion rate, which supports our hypothesis.
As the number of required action steps of tasks increases, the success rates of all the methods decrease as expected.
By considering action knowledge~(i.e., preconditions and effects), \tpvqa ~and EffectVQA are significantly better than others, especially the ones~(PaLMEVQA and SuccessVQA) that only query about actions by their names.   
We can also tell that methods additionally considering action affordances~(\tpvqa ~and PaLMEVQA) perform better than the methods that only detect action failures~(EffectVQA and SuccessVQA).  
% \begin{figure*}
% \begin{center}
%     \includegraphics[width=0.9\textwidth]{figures/side.pdf}
%     \caption{
%     }
% \end{center}
% \end{figure*}

Another interesting finding is that TP, a baseline that does not include any perception, produces a higher success rate than PaLMEVQA and SuccessVQA, which are two other baselines that are capable of interacting with VLMs.
\begin{table}
    \centering
    \begin{tabular}{ l  r r r}
    \toprule 
    \multirow{2}{*}{Query} & \multicolumn{3}{c}{Accuracy} \\
    \cmidrule(lr){2-4}
                               & clean\_dishes & serve\_breakfast & eat\_apple \\
    \midrule
    Action Pres.             & \textbf{0.63}        & \textbf{0.53}                 & \textbf{0.70}        \\
    Is $<$action$>$ possible?               & 0.58        & 0.33                 & 0.43        \\
    \midrule
    Action Effs.             & \textbf{0.79}        & \textbf{0.60}                 & \textbf{0.71}        \\
    Was $<$action$>$ successful?               & 0.45        & 0.30                 & 0.47\\
    \bottomrule
    \end{tabular}
\caption{VQA accuracies on different querying strategies.}
\vspace{-2em}
\label{tab:side_result}
\end{table}
That is because false positives and false negatives from VLMs will greatly impact the plan execution, easily leading the robot to failure cases.
TABLE~\ref{tab:side_result} shows that when querying about action knowledge, the VLM is more accurate in failure detection and affordance prediction.
If directly querying about action names, prediction accuracies for most of the actions are lower than a random guess, which is why SuccessVQA and PaLMEVQA perform poorly. 
A straightforward example is when detecting if the robot has successfully washed the plate, instead of asking \textit{``Was wash plate successful?''}, \tpvqa ~will query \textit{``Is the plate clean?''}.
We observe that the latter type of queries~(ours) is easier for VLMs to understand.

\subsection{Real-Robot Deployment}
\begin{figure}[h!]
\begin{center}
    \includegraphics[width=0.43\textwidth]{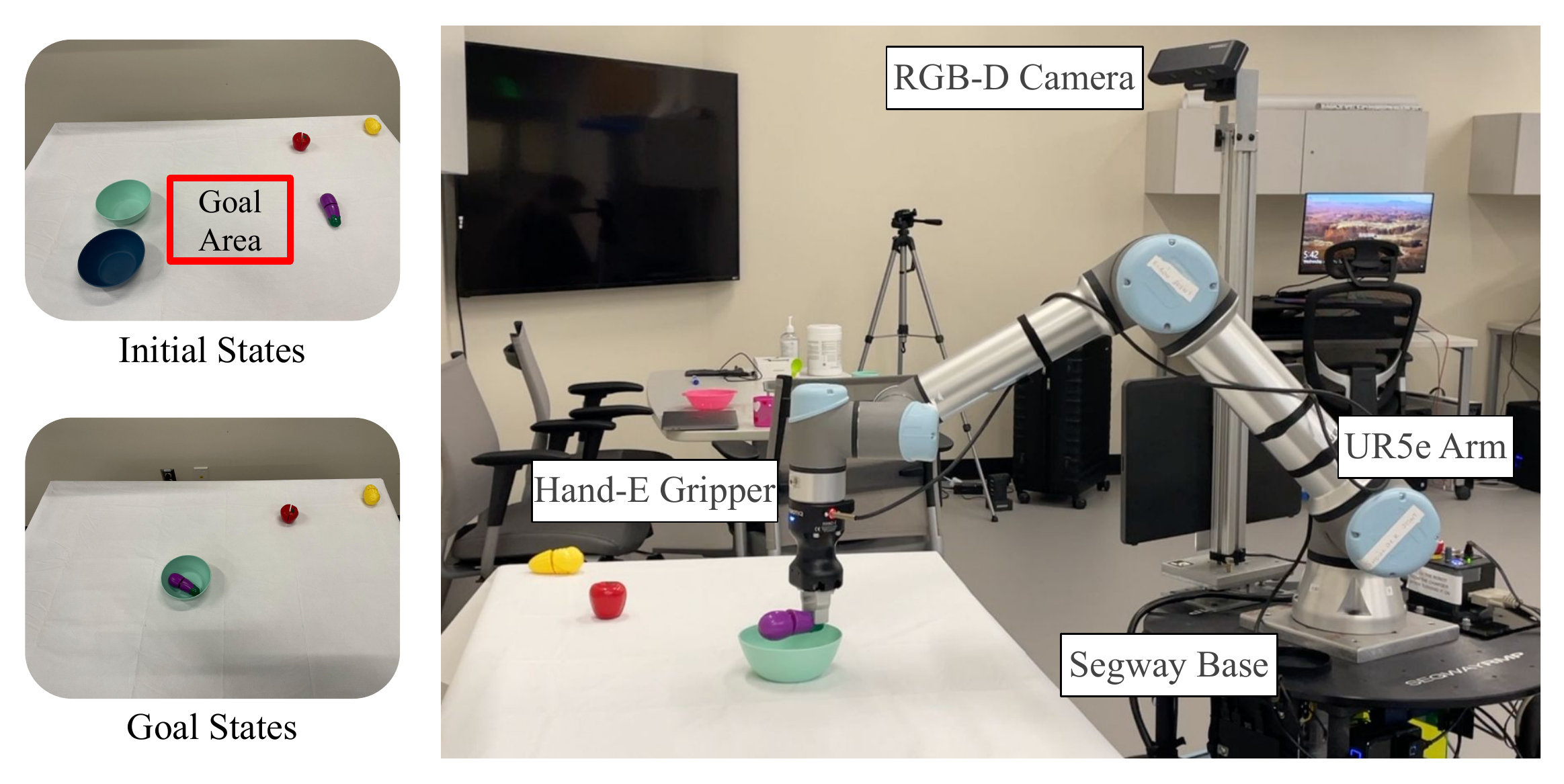}
    \caption{Real robot demonstration for \tpvqa.
    }
    \vspace{-2em}
    \label{fig:real_short}
\end{center}
\end{figure}

We also deployed \tpvqa ~on real robot hardware to perform object rearrangement tasks~(Fig.~\ref{fig:real_short}), where the goal is to ``collect'' toys using a container and place them in the red area.
Our real-robot setup includes a UR5e Arm with a Hand-E gripper mounted on a Segway base, and an overhead RGB-D camera (relatively fixed to the robot) for perception.
Please refer to Appendix for more details.

\section{Conclusion and Future Work}
In this paper, we investigate robot classical planning with pre-trained VLMs.
We propose \tpvqa ~that triggers re-planning using precondition checking and re-execution using effect monitoring.
By doing a set of experiments on robots working on everyday tasks, we demonstrate that \tpvqa ~is able to provide more successful plan executions than baselines.
For future work, we would like to evaluate our method using more tasks, potentially those from existing benchmarks such as ActivityPrograms~\cite{puig2018virtualhome}.
In addition to using images generated by diffusion models for evaluation, it might be possible to use datasets such as EGO4D~\cite{grauman2022ego4d}.
To improve the overall task completion rate, one way is to collect task-specific data and finetune the pre-trained model.
It will be also interesting to develop methods that can handle inconsistent inferences from VLMs.

\bibliographystyle{IEEEtran}
\bibliography{ref}
\begin{figure*}
\begin{center}
    \includegraphics[width=\textwidth]{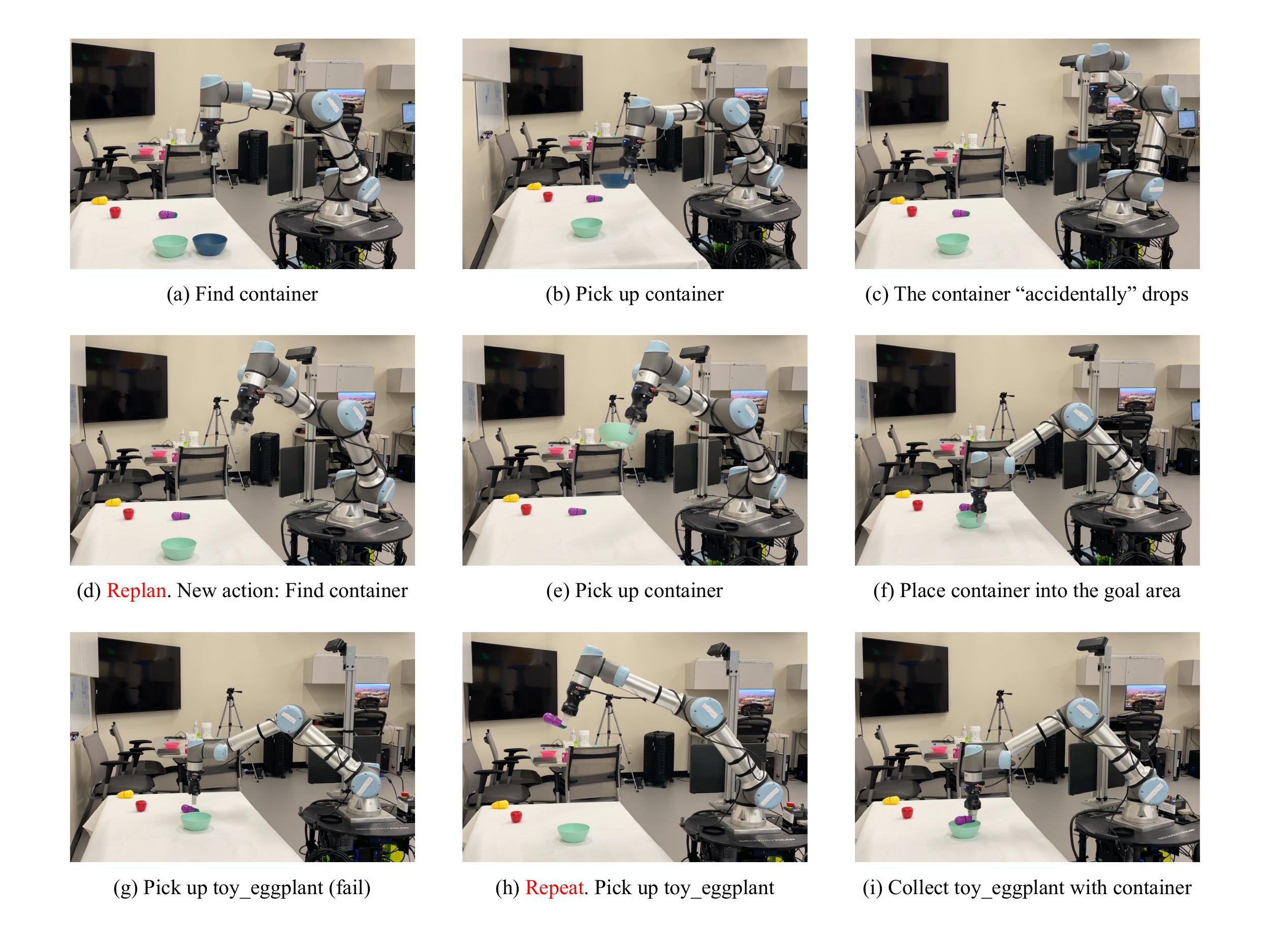}
    \caption{Screenshots showing the full demonstration trial of \tpvqa ~as applied to a real robot.
    }
\label{fig:real_long}
\end{center}
\end{figure*}
\newpage

\section*{APPENDIX}
Fig.~\ref{fig:real_long} shows a sequence of screenshots of a real robot using \tpvqa ~on object rearrangement tasks.
The goal is to ``collect'' toys using a container and place them in the middle of the table~(i.e., goal area).
We assume that the robot has a predefined set of skills, including \texttt{pick}, \texttt{place}, and \texttt{find}.
\texttt{Pick} and \texttt{place} actions are implemented using GG-CNN~\cite{morrison2018closing}, and \texttt{find} action simply uses base rotation for capturing tabletop images from different angles.

Given the task description, the robot first decided to execute ``find container'' and ``pick up container''.
These two actions were successfully executed as shown in Fig.~\ref{fig:real_long}(a),~\ref{fig:real_long}(b).
When the robot was preparing the next action~(i.e., ``Place container into the goal area''), the blue container accidentally dropped from the robot's gripper to the ground~(Fig.~\ref{fig:real_long}(c)).
Instead of directly executing the next action, \tpvqa ~enabled the robot to check preconditions by querying the VLM \textit{``Is the container in a robot's hand?''}
After receiving negative feedback from the VLM, \tpvqa ~updated the world state by removing \texttt{in\_hand(container)} and called the planner to generate a new plan that started the task again by finding another container~(Fig.~\ref{fig:real_long}(d)).
Then the robot picked up the cyan container and placed it in the middle of the table as shown in Fig.~\ref{fig:real_long}(e),~\ref{fig:real_long}(f).
The subsequent actions in the plan were to find and pick up a toy, but the \texttt{pick} action failed~(Fig.~\ref{fig:real_long}(g)).
\tpvqa ~managed to detect the failure by querying 1) \textit{``Is there a toy\_eggplant on the table?''}, and 2) \textit{``Is the toy\_eggplant in a robot's hand?''}, and receiving Yes and No answers respectively.
As a result, our system suggested the robot repeat the \texttt{pick} action again~(Fig.~\ref{fig:real_long}(h)).
Finally, the robot successfully collected the toy by putting it into the cyan container that was previously placed in the goal area~(Fig.~\ref{fig:real_long}(i)).
\end{document}